\pdfoutput=1
\documentclass[journal]{vgtc}              

\onlineid{2215}

\vgtccategory{Research}

\vgtcpapertype{algorithm/technique}

\title{Explicit 3D Gaussian Models for Fast, Mesh-Free Compression of Structured and Unstructured Volumes}

\newcommand{\familys}[1]{directly queryable models}
\newcommand{\fams}[1]{DQM}

\author{%
  Landon Dyken,
  Sharmistha Chakrabarti, 
  Nathan Debardeleben,
  Steve Petruzza,
  Qi Wu,
  Will Usher,
  Sidharth Kumar
}

\abstract{%
    Recent work has shown that implicit neural representations (INRs) can be trained to effectively compress structured and unstructured volume data, allowing for direct data querying with a reduced memory footprint. However, as existing INRs for unstructured volumes do not encode geometry, they require partial mesh storage for later sampling, limiting achievable compression. At the same time, novel view synthesis methods have shown that explicit collections of 3D Gaussians can be used to accurately visualize volume data. In this work, we introduce an explicit model for volume data compression based on 3D Gaussian primitives. We reinterpret collections of 3D Gaussians as an explicit representation of a volume’s scalar field and use a sampling strategy that reconstructs scalar values at spatial locations through weighted aggregation of intersecting Gaussians. We develop optimized CUDA-accelerated pipelines for structured and unstructured model sampling, loss functions that encourage accurate domain coverage of our models, and a novel sampling-error based densification strategy. Our explicit formulation naturally encodes domain geometry, eliminating the need for mesh storage in unstructured volumes and introducing significantly higher compression opportunities. Compared to existing INRs, we demonstrate that our explicit model achieves significant training speed and reconstruction quality improvements on both structured and unstructured volumes, while enabling compression ratios much higher than previous work for unstructured volumes.
}


\teaser{
  \centering
  \includegraphics[width=\linewidth, alt={Asteroid impact.}]{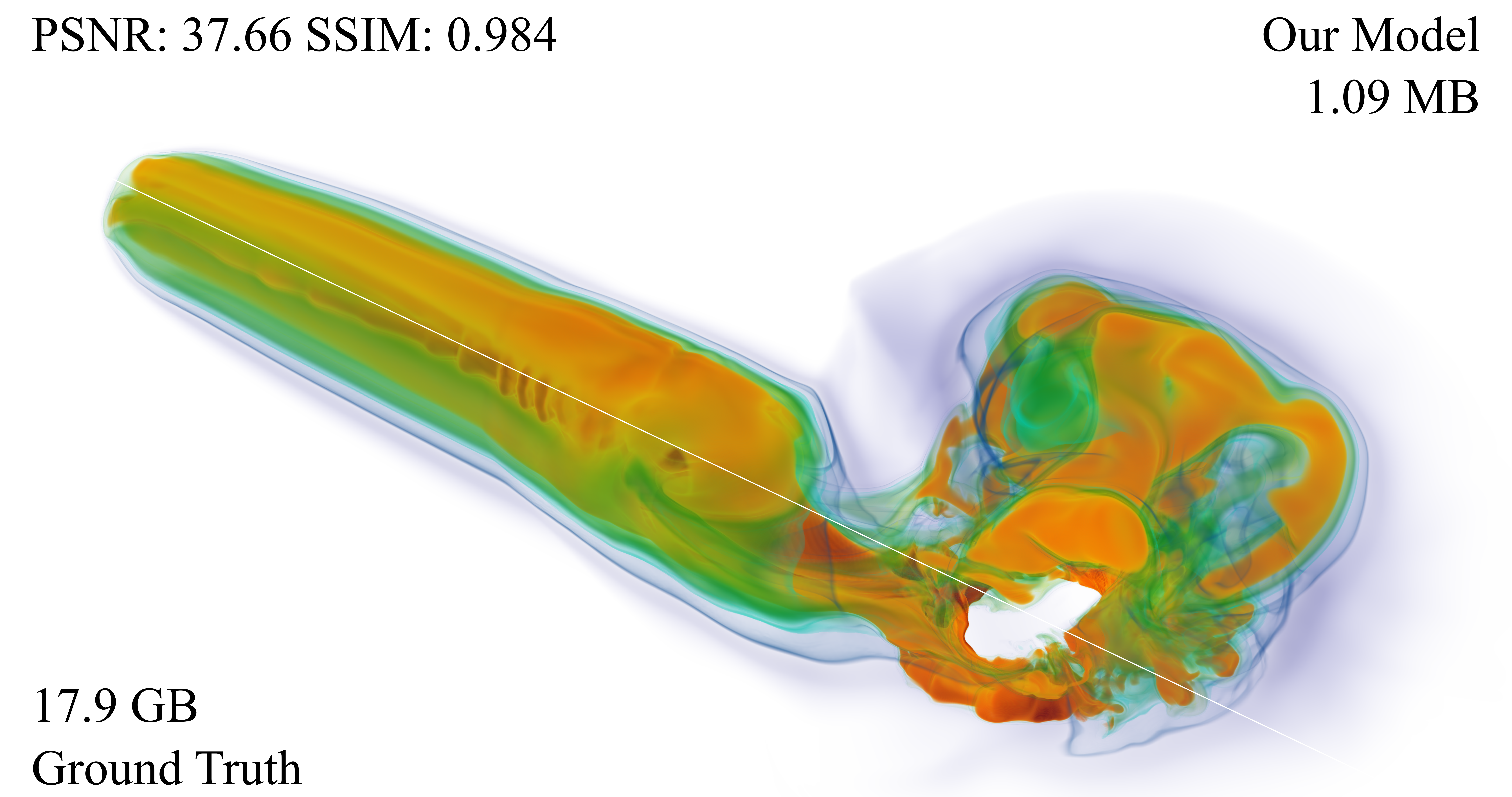}
  \vspace{-1em}
  \caption{%
  	Volume rendering of the 17.9 GB unstructured Impact dataset with a custom transfer function using the ground truth and our 16,384$\times$ compressed model, at 4K samples per pixel. Our method encodes both scalar field and domain geometry using 3D Gaussians, removing the need to store mesh geometry, which limits existing methods to a maximum real compression ratio of $18.7\times$ on this dataset. This model reaches a volumetric PSNR of 50.33 dB.
  }
  \label{fig:teaser}
}

\graphicspath{{figs/}{figures/}{pictures/}{images/}{./}} 

\usepackage{tabu}                      
\usepackage{booktabs}                  
\usepackage{lipsum}                    
\usepackage{mwe}                       
\usepackage{booktabs, multirow}
\usepackage{amsmath}
\usepackage{amssymb} 

\usepackage{mathptmx}                  
\usepackage{siunitx}
\begin{document}


\maketitle

\section{Introduction}
\label{sec:intro}
Modern HPC systems have enabled scientific simulations to generate massive volumetric datasets, regularly reaching gigabytes to terabytes for storing a single simulation timestep. While simulations use adaptive mesh refinement (AMR)~\cite{anderson2024fun3d, flash_2014, lava_2016,BERGER198964, arndtDealIIFiniteElement2021, bursteddeP4estScalableAlgorithms2011} or unstructured meshes~\cite{anderson2024fun3d, economonSU2OpenSourceSuite2016, offermansAdaptiveMeshRefinement2020, palaciosStanfordUniversityUnstructured2013} to more efficiently utilize limited memory budgets, output volumes remain large. Effectively using these datasets for end user tasks is difficult, as they exceed the memory of client systems, necessitating further use of HPC resources. While traditional compression methods can greatly reduce data size, they require decompressing some or all of the data before it can be used. This leads to latency and added memory usage, which prohibit the use of the compressed model for many tasks. 

In contrast, {\familys{}} offer compression that does not require decompression before sampling the data. A family of methods that has become popular for this use case is the creation of scene representation networks (SRNs), also called implicit neural representations (INRs)~\cite{Weiss_2022, wu2024interactivevolumevisualization, lu_neurcomp_2021,wurster23_APMGSRN}, which approximate volume data with neural networks that learn an implicit function from input 3D positions to output scalar values. While most approaches for volume visualization have targeted only structured grid data, there have also been extensions to unstructured volumes~\cite{liu24_uginr, son2025_mcinr}, by clustering data points and training separate SRNs to represent each cluster. This strategy results in a compressed model of the scalar field contained in an unstructured mesh, but does not encode any information about the geometry of the mesh itself. As a result, some uncompressed mesh geometry must be stored alongside the model for downstream tasks, setting a hard cap on achievable compression ratios. For the datasets in our evaluation, this storage limits real effective compression ratios to between $7$--$24\times$. 

At the same time, a separate family of methods has explored applying novel view synthesis (NVS) techniques to volume visualization~\cite{yao_revolve_2025, yao_visnerf_2025, nli4volvis_2025, ivrgs_2025, veg_2025} to render datasets within small memory and compute footprints. Instead of creating a compressed representation of a volume dataset directly, these methods train a model from rendered images of a dataset, producing a representation that can infer renderings with unseen camera viewpoints, lighting parameters, and even transfer functions applied. Generally, while these models can produce accurate images, their applications are constrained to a specific rendering method. Additionally, training times for novel view synthesis models are typically much longer than those of SRNs~\cite{wu2024interactivevolumevisualization, veg_2025, yao_revolve_2025}. 

In this work, we take inspiration from \textit{Volume Encoding Gaussians (VEG)}~\cite{veg_2025}, a novel view synthesis method in which 3D Gaussian splatting (3DGS)~\cite{3dgs} was improved for volume visualization by replacing attached color properties with scalar values per Gaussian. We adapt this representation for direct compression of volume data by creating an inference pipeline to reconstruct volume samples in 3D space, rather than images in 2D space. Output is computed through averaging the values of 3D Gaussians intersecting a sample location, effectively treating the collection of Gaussians as a representation of the volume's underlying scalar field. By using fewer Gaussians than data points in the ground truth, our models can represent a dataset using a fraction of the original memory, while supporting direct sampling as if they were the original data. We call the resulting representation \textit{Volume Encoding Gaussians with Geometry-Integration and Explicit Sampling} (VEGGIES). 

Our method differs from existing learning-based compressors in that we build an \textit{explicit} model, rather than an implicitly defined neural network, which has several advantages.  First, using our CUDA-accelerated sampling algorithms, our training process is faster than every INR-based method we tested, and orders of magnitude faster than the most accurate. Second, our model has completely dynamic memory allocation; Gaussians can be added or removed in order to best adapt to the compression goal and dataset being represented. We utilize this by presenting a novel densification strategy which places new primitives at world positions with high sampling error during training. This technique improves reconstruction quality greatly, leading to our models reaching higher reconstruction quality than all tested INRs across datasets and compression ratios. Finally, our model can directly define the domain of a scalar field with the geometry of 3D Gaussians. By introducing some lossiness in domain representation, our method is able to create arbitrarily small compressed models.

Our contributions are as follows: 
\begin{enumerate}
    \item An \textit{explicit} trained model for volume data compression with CUDA-accelerated sampling for structured and unstructured data. This model reconstructs structured volumes with an average $7.02$\,dB higher PSNR than the fastest structured INR, while also training faster on all twelve benchmarks tested (geometric mean $1.42\times$).
    \item A sampling-error based densification strategy that improves reconstruction quality. With this method, our model outperforms the highest quality INR on every benchmark that method was able to complete, while training $2{,}800\times$ faster (geometric means of 11.7 seconds against 9.2 hours).
    \item The first model for unstructured volume compression that directly encodes the volume domain. At the same model sizes, this method improves PSNR by at least $6.20$\,dB on every dataset UGINR could run on, trains $5.2\times$ faster (geometric mean), and removes the real compression ratio ceiling from exterior surface storage, reaching $18\times$ higher effective compression ratios across our benchmarks (geometric mean).
\end{enumerate}
 



\section{Related Work}
\label{sec:related_work}
We briefly review prior work in Gaussian models for volume data in Section~\ref{sec:related_work_gmm}. Although our work does not use a neural network, it targets the same use cases as work on neural representations for volume data, so we review this in Section~\ref{sec:related_work-neural}. Finally, Section~\ref{sec:related_work-image} reviews prior work on NVS techniques for volume visualization, as these provide inspiration for our technique.

\subsection{Gaussian Representations of Scalar Fields}
\label{sec:related_work_gmm}
Gaussians have a long history as a compact representation of volume data. In visualization, Gaussian mixture models (GMMs) fit by expectation-maximization have mainly represented value distributions, per voxel~\cite{Liu2012GMM} or per block with a spatial component~\cite{Wang2017}. Radial basis function (RBF) encodings instead approximate the field with truncated spherical~\cite{Jang2004,Weiler2005}, ellipsoidal~\cite{Jang2006}, or axis-aligned anisotropic~\cite{Juba2007} Gaussians, rendered directly from the encoding. RBF parameters are fit in stages: center placement by clustering or greedy error-driven selection, width estimation by nonlinear optimization~\cite{Jang2004} or a closed-form maximum-likelihood estimate~\cite{Juba2007}, and a least-squares solve for weights at the input data points. 

VEGGIES differs from this line of work in three ways. First, we optimize all parameters jointly by gradient descent through differentiable GPU sampling. Second, RBF encodings evaluate an additive expansion, $f(p) = w_0 + \sum_i w_i G_i(p)$, which assigns a value everywhere in space; Equation~\ref{eq:scalar} instead carries a separate learned weight and scalar value per Gaussian and returns a weighted average of values normalized by total weighted influence, defined only where that influence exceeds a threshold $\mathrm{T}$ (Section~\ref{sec:estimate_samples}). Third, prior encodings leave the volume's domain implicit, relying on later processing like value clipping and transfer function selection to hide the background outside the data~\cite{Jang2004}. In contrast, our formulation, along with our false negative (Section~\ref{sec:continuity}) and false positive (Section~\ref{sec:method_fp}) losses, train the model's support to accurately match an unstructured volume's domain.

\subsection{Neural Representations for Volume Data}
\label{sec:related_work-neural}
Prior work has proven neural representations greatly effective for volume compression and visualization tasks, as shown in the recent survey by Wang and Han~\cite{wang_dl4scivis_2023}. These representations use neural networks to estimate a volume's scalar field, which can allow for faster sampling, lowered memory footprint, and more interactive rendering than usage of the original volume.
Early work by Griffin et al.~\cite{Jain2017CompressedVR} proposed a deep convolutional autoencoder network for compressed rendering of multivariate time-varying volume data. 
After the success of SIREN's~\cite{sitzmann_siren_2020} implicit neural representation for general images and 3D shapes, Lu et al. created \textit{NeurComp}~\cite{lu_neurcomp_2021} to adapt the method for volume data compression. To reduce computational cost, subsequent work introduced feature-grid INRs, which enable shallower networks by using learned spatial encodings. Weiss et al. developed fV-SRN~\cite{Weiss_2022} using dense-grid encoding~\cite{takihawa_dense_2021} to support interactive rendering of volume data with their neural representation. Wu et al. then utilized multi-resolution hash encoding~\cite{muller_multi_2022} and algorithmic improvements to rendering to further build on fV-SRN's performance with InstantVNR~\cite{wu2024interactivevolumevisualization}. Wurster et al.~\cite{wurster23_APMGSRN} created a method using adaptively placed multi-grids for their SRN (AMGSRN), then improved its speed and memory footprint with fused CUDA kernels and compression-aware training~\cite{wurster25_AMGSRN}. Han et al.~\cite{han_moeinr_2026} used a mixture-of-experts framework to divide spatiotemporal fields for time-varying volume compression. While the methods reviewed so far have only been applied to structured volumes, there is recent work in extending neural representations to unstructured volume data. Liu et al.~\cite{liu24_uginr} presented UGINR, which used k-means clustering to separate groups of data points in an unstructured volume, then compressed each with a separate network. Son et al.~\cite{son2025_mcinr} then proposed MC-INR, which extended this method through meta-learning across clusters. 

While our method produces a learned model for volume compression, it utilizes an explicit set of Gaussian primitives rather than a neural representation. Our model controls memory allocation in space at the level of individual primitives, via adaptive densification of 3D Gaussians. This benefits datasets with empty space or varying feature density, and maps especially well to the complexity of unstructured volumes. Our method also skips the expensive k-means clustering steps of MC-INR and UGINR, and provides the opportunity to encode the volume domain, whose absence greatly limits compression ratios in their cases.

\subsection{Novel View Synthesis for Volume Visualization}
\label{sec:related_work-image}
As they are trained on sets of images rather than data directly, novel view synthesis methods have the capability to render volumes while having compute and memory costs not tied to the data size. Early work by Berger et al.~\cite{berger_generative_2019} showed that pre-trained generative adversarial networks can synthesize volume rendering and transfer function exploration. Yao et al.~\cite{yao_revolve_2025} proposed a hybrid approach where a network is trained from a small number of rendered images to reconstruct pre-shaded volumetric scenes. VisNeRF~\cite{yao_visnerf_2025} presented a radiance field representation that allows for NVS of volume data with interactive rendering parameter control. Because of its popularity for general NVS, recent work has explored 3D Gaussian splatting (3DGS)~\cite{3dgs} for volume visualization. This work involves representing a 3D scene as a set of 3 dimensional Gaussians with colors and opacities, that can then be projected into 2D and blended to render an image. As this rendering is differentiable, the set of Gaussians can be optimized from training images to reconstruct the scene accurately, while the explicit model of 3D Gaussians allows for fast rendering. Han et al.~\cite{han_isosurface_2025} recently used distributed 3DGS~\cite{zhao2024scaling3dgaussiansplatting} for visualization of large scientific datasets. iVR-GS~\cite{ivrgs_2025} adapted 3DGS to volume visualization by implementing Blinn-Phong lighting, editable Gaussian colors, and composed sets of Gaussians to represent a volumetric scene at multiple transfer functions. Dyken et al.~\cite{veg_2025} built on this by introducing volume encoding Gaussians (VEG), which achieve better performance on transfer functions unseen during training by removing color and opacity per Gaussian, replacing them with scalar values that learn the volume more directly. 

Our method utilizes the VEG representation, but applies it to volume compression instead of novel view synthesis. Rather than only rendering in a specific configuration, our models can be used for any sampling-dependent task. We also find our world-space training pipeline to be much faster than image-space training, with the original VEG benchmarks~\cite{veg_2025} taking 10--20 minutes on the Chameleon, RBL, and Mito datasets compared to less than a minute for ours. To avoid ambiguity, we hereafter refer to the original rendering-oriented formulation as image-space VEG (I-VEG).  
\section{Preliminaries}
\label{sec:prelim}
In this section, we first give background on volume data types (Section~\ref{sec:volume_types}), then review the VEG representation (Section~\ref{sec:veg}), and how we adapt it to volume sampling. In Appendix~\ref{sec:veg_failure}, we additionally present a demonstration of how I-VEG models~\cite{veg_2025} fail to generalize to volume reconstruction. 

\begin{figure}[t]
    \centering
    \includegraphics[width=\linewidth]{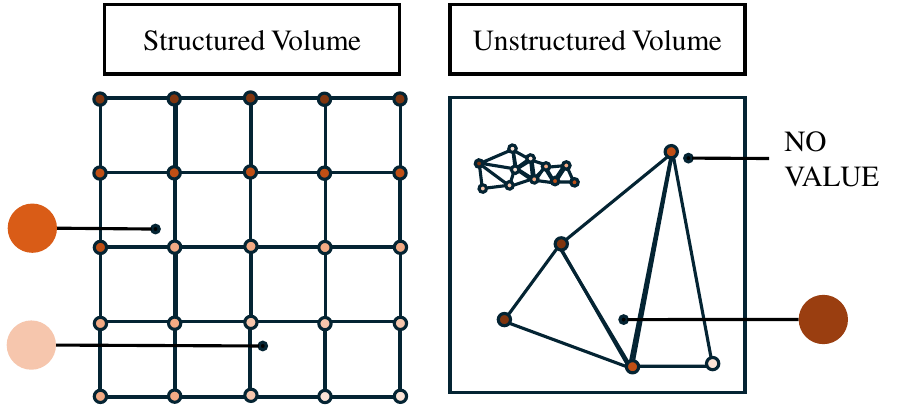}
    \caption{Visualization of sampling for structured (left) and unstructured (right) volumes. Structured volumes define scalar fields over a regular grid, while unstructured volumes must define an explicit geometry.}
    \label{fig:volumes}
    \vspace{-1.5em}
\end{figure}
\subsection{Volume Types}
\label{sec:volume_types}
A structured volume is defined by a set of scalar values on a regular grid, with neighboring points assumed to be connected by cells. Any point in space can be interpolated from surrounding grid points. The geometry of a structured volume is implicit; only the data values need to be stored. In contrast, unstructured volume data requires explicit storage of the geometry over which the scalar field is defined. Given a position in space, a test is performed to determine which cell in the volume, if any, contains it, before interpolating between that cell's vertices to find the scalar value at that position. A comparison of sampling structured and unstructured volume data is shown in Figure~\ref{fig:volumes}. 

Both structured and unstructured volume data can be defined as functions that map positions in a subset of three-dimensional space to values in a scalar field, i.e., $\Phi: \Omega \subset \mathbb{R}^3 \rightarrow \mathbb{R}, (x,y,z)\mapsto v=\Phi(x,y,z)
$. They differ in that, while the domain of a structured volumetric function is easily defined by its origin and bounds, the domain of an unstructured volumetric function has no set structure; the function can be defined over any region matching the unstructured volume's geometry. In addition, the complexity of this function is not uniform with respect to spatial area, as points in an unstructured volume can have arbitrary spacing. 

\subsection{Volume Encoding Gaussians}
\label{sec:veg}
Volume encoding Gaussians are 3D Gaussian distributions that differ from traditional 3D Gaussians~\cite{3dgs} by carrying trainable scalar values and weights rather than colors and opacities. I-VEG~\cite{veg_2025} additionally attach lighting parameters, apply a transfer function per Gaussian, and splat the result into 2D image space with front-to-back $\alpha$-blending, so the model is only ever evaluated through rendered images. We instead target volume reconstruction, and introduce VEGGIES, which are trained in world space and evaluated directly at 3D sample positions. Each VEGGIES primitive is parameterized by its mean position ($\mu_i$), scaling and rotation matrices defining its 3D covariance ($S_i$, $R_i$), scalar value ($v_i$), and weight ($w_i$).

To evaluate our models at a point, we use normalized Gaussian kernel regression: given the $N$ Gaussians whose influence reaches a point $p=(x, y, z)$ (with the cutoff and stability floor defined in Section~\ref{sec:estimate_samples}), we perform estimation of the value of the scalar function $\phi$ at that point as the weighted average:
\begin{equation}
\Phi(p) = \frac{\sum_{i=1}^{N}{w_i\cdot v_i \cdot G_i(p)}}{\sum_{i=1}^{N}w_i \cdot G_i(p)}
\label{eq:scalar}
\end{equation}
where each $G_i$ is the three-dimensional Gaussian kernel defined as:
\begin{equation}
    G_i(p) = e^{-\frac{1}{2}(p-\mu_i)^\top R_{i}^{\top}S_i^{-2}R_i(p-\mu_i)}
\end{equation}
with $R_{i}^{\top}S_i^{-2}R_i$ being the inverse of the i-th Gaussian's 3D covariance matrix. We multiply by the learnable weight parameter $w_i$ to provide a distance-independent way for Gaussians' influence to be scaled up or down. Similarly to the weight parameter in previous work~\cite{veg_2025}, this provides the training process direct control to reduce the effect of Gaussians that contribute negatively to reconstruction, as well as increase the importance of useful Gaussians. 

\begin{figure*}[t]
    \centering
    \includegraphics[width=\linewidth]{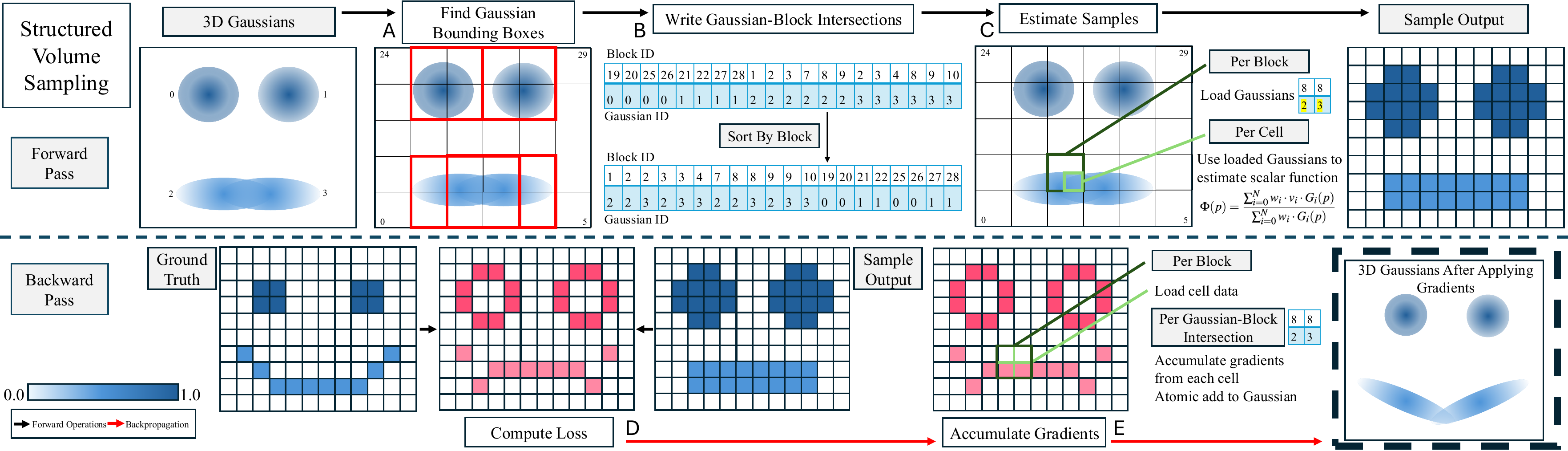}
   \caption{Our sampling algorithm on a $10\times12$ structured dataset with $2\times2$ blocks, showing the forward (A--C) and backward (D--E) passes. (A) AABBs are computed for each Gaussian in block units. (B) Each Gaussian-block intersection is written as a [Block ID, Gaussian ID] pair and sorted by block. (C) Each block loads its intersected Gaussians and each cell evaluates Equation~\ref{eq:scalar}. (D) Loss is computed against the ground truth and differentiated with respect to each sample. (E) The gradient kernel reuses the sorted list from (C): each block loads its cell data, and each Gaussian sums the partial gradients from the cells it influenced before one atomic add to its global gradient. This figure shows 0-valued cells as empty for simplicity. 
    }
    \label{fig:structured_method}
    \vspace{-1em}
\end{figure*}

\section{Structured Volume Reconstruction}
\label{sec:structured_method}
We first present our structured pipeline, which adapts the tile-based 2D rasterizer of Kerbl and Kopanas et al.~\cite{3dgs} to voxel sampling on a regular 3D grid, computing scalar values at sample locations with Equation~\ref{eq:scalar} (Section~\ref{sec:structured_implementation}), then describe how VEGGIES models are trained using this pipeline (Section~\ref{sec:structured_training}).

\subsection{Structured Volume Implementation}
\label{sec:structured_implementation}
Figure~\ref{fig:structured_method} illustrates our pipeline's forward pass, which finds Gaussian bounding boxes (A), writes Gaussian-block intersections (B), and estimates samples (C), along with the custom backward pass that accumulates Gaussian gradients from the loss (D--E).

\subsubsection{Finding Gaussian Bounding Boxes}
\label{sec:gaussian_bounding}
To reconstruct samples for each cell, the sampling kernel needs to sum all relevant Gaussian-cell intersections. 
Because the influence of a Gaussian is effectively zero for most of the volume, we create Gaussian bounding boxes for the ellipsoid around the Gaussian's mean where its influence is greater than the cutoff ($\tau_1$), and use these to write out intersections. For our evaluation, we set $\tau_1=0.0001$ for all models as a solid tradeoff between reducing unnecessary intersections and maintaining accuracy. Because neighboring cells are likely to intersect the same Gaussians, per-cell intersection lists can be replaced with intersection lists for blocks of neighboring cells without drastically increasing the sizes of the lists, so axis-aligned bounding boxes (AABBs) are computed at the block level.


\subsubsection{Writing Gaussian-Block Intersections}
\label{sec:gaussian_intersections}
Next, we write out the Gaussian and block IDs for each intersection using the computed AABBs. Unlike the tile-based rasterizer from 3DGS~\cite{3dgs}, which parallelizes this step over Gaussians, our method parallelizes this step over total intersections, using a binary search and indexing to find which Gaussian and block it corresponds to respectively. Individual Gaussians intersect many more cells in 3D sampling than pixels in 2D rasterization, with far greater variance in count, leading to poor workload balance with their method. After writing intersections, we sort the Gaussian and block IDs by block ID and write out the indices of each block's list of intersections, giving the input needed for the sampling kernel. 

\subsubsection{Estimating Samples}
\label{sec:estimate_samples}
The final kernel in our forward algorithm estimates the samples for each cell $c$. This is done by launching a thread for each cell in the grid, organized into thread blocks according to the blocks of the volume. Each thread block cooperatively loads a batch of Gaussians from its intersection list into shared memory. Threads use these to compute contributions from all Gaussians whose influences are above the cutoff ($w_i \cdot G_i(p_c) > \tau_1$) and add these to their cells' accumulated influence ($I_c \mathrel{+}=w_i \cdot G_i(p_c)$) and accumulated weighted value ($V_c \mathrel{+}=v_i \cdot w_i \cdot G_i(p_c)$), where $p_c$ is the sample location. Once all Gaussians in a block have been processed, each thread writes its sample output as $\hat{\Phi}_c=\frac{V_c}{\max(I_c, \tau_2)}$, where the floor $\tau_2$ ensures numerical stability for low $I_c$. For our evaluation, we set $\tau_2=0.01$ for all models. A position can thus be given a value by the model exactly when at least one Gaussian has influence on it above $\tau_1$. However, to later improve domain encoding accuracy, we introduce a separate threshold $\mathrm{T} > \tau_1$, where a point is only said to be defined by a VEGGIES model if $I_c > \mathrm{T}$. For our evaluation, we set $\mathrm{T}=0.002$, except for models that encode an unstructured volume's domain (Section~\ref{sec:method_fp}), where $\mathrm{T}=0.01$.

\subsubsection{Accumulating Gaussian Gradients}
\label{sec:accumulate_gradients}
PyTorch supplies partial derivatives of the loss with respect to each sample, which our backward kernel propagates to the Gaussians that influenced it. Launching one thread per cell, as in the forward pass, causes heavy atomic contention, since every cell a Gaussian touches adds its partials to that Gaussian's global gradient. Instead, we reuse the sorted intersection list from the forward pass to launch one thread per Gaussian-block intersection, grouped by block. Each block loads its cell data ($I_c$, $\hat{\Phi}_c$, $\frac{dL}{dI_c}$, $\frac{dL}{d\hat{\Phi}_c}$) into shared memory, and each thread sums the gradients for its Gaussian over those cells before a single atomic add, reducing atomic operations by a factor of the block size. This mirrors recent per-splat gradient parallelization for the 3DGS rasterizer~\cite{mallick_taminggs_2024,hahlbohm2026fastergs}. Across the pipeline, a block size of 32 ($4\times4\times2$ over $x$, $y$, $z$) performs best.

\subsection{Structured Volume Training}
\label{sec:structured_training}
With our fast differentiable sampling engine, training proceeds through successive iterations of sampling a set of VEGGIES, computing loss against ground truth samples, then backpropagating gradients to the parameters of each Gaussian ($\mu_i$, $S_i$, $R_i$, $v_i$, $w_i$). We use sigmoid activation functions for $v_i$ and $w_i$ to constrain them within $(0, 1)$ for stability, a normalization activation for $R_i$, and an exponential activation for $S_i$. We apply a soft norm cap to scales to reduce workload imbalance that occurs with Gaussians of wildly different sizes. We use L2 reconstruction loss between model and ground truth samples. To initialize our models, we use the strategy of Dyken et al.~\cite{veg_2025}, converting a portion of a volume's vertices to isotropic VEGGIES with the same scalar values, initial weights of 0.1, and scales dependent on the density of surrounding points. In Section~\ref{sec:densification}, we discuss methods for density control of VEGGIES during training, and in Section~\ref{sec:continuity} we discuss how to ensure VEGGIES models cover a volume's full domain.  


\begin{figure}[t]
    \centering
    \includegraphics[width=\linewidth]{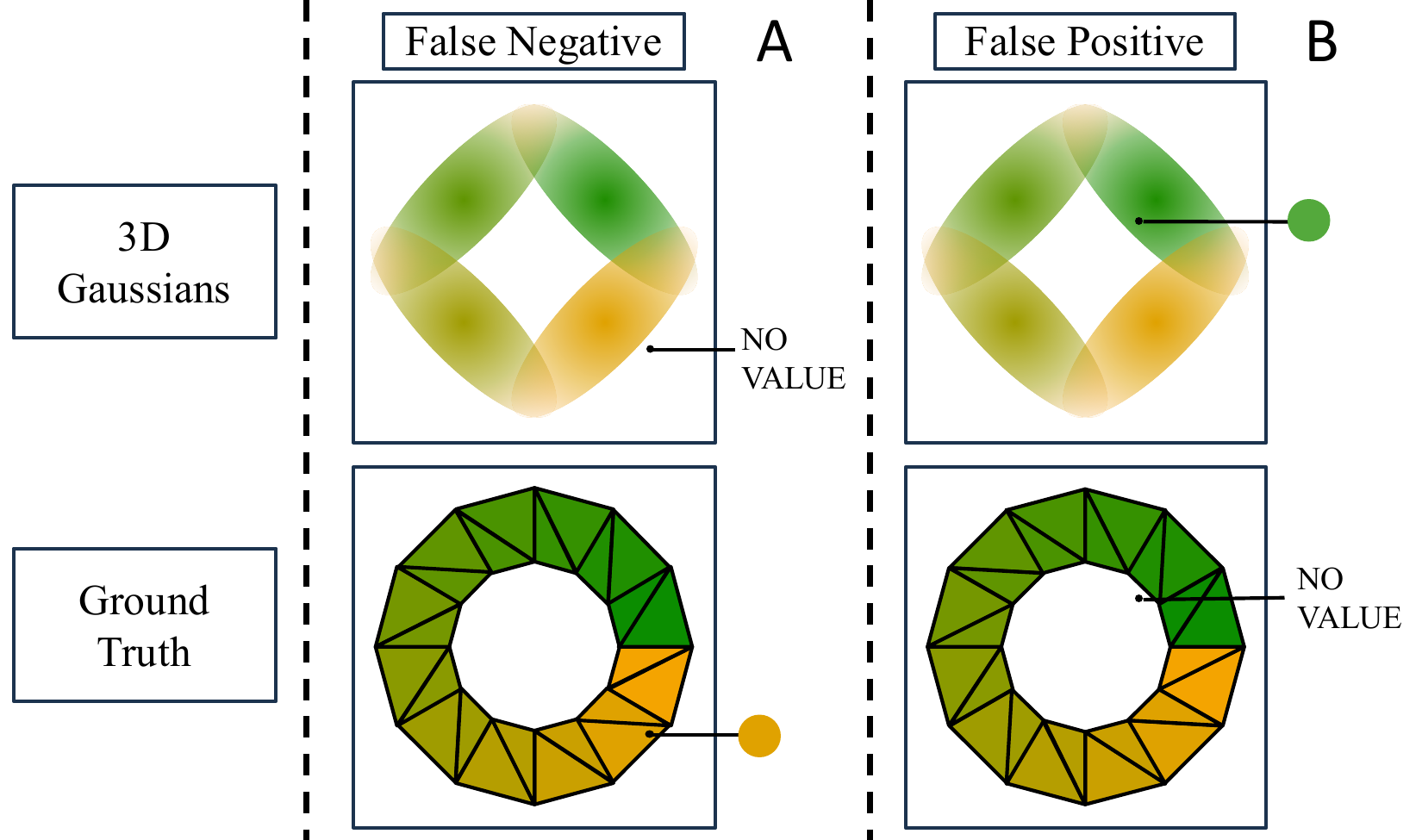}
    \caption{Illustration of how we define false negative (A) and false positive (B) samples in a VEGGIES model for an example unstructured volume. (A) False negative samples are undefined in the model, even though they are given a value in the original volume. (B) False positive samples are defined by the model, but are undefined in the original volume. We introduce loss terms in our training process to minimize the occurrence of both of these model failures.}
    \label{fig:fpfn}
    \vspace{-1.5em}
\end{figure}

\begin{figure*}[t]
    \centering
    \includegraphics[width=\linewidth]{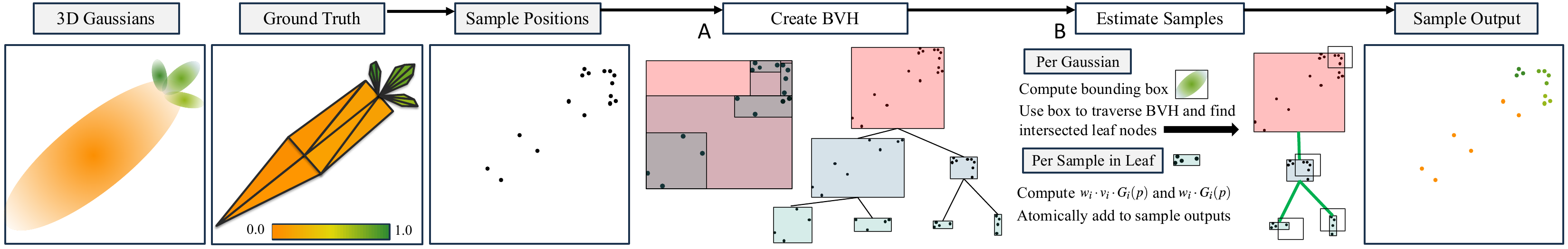}
    \caption{An illustration of our sampling algorithm on a 16 cell unstructured dataset, where a sample is taken at each cell center. (A) We first create a BVH containing the given set of positions to sample. This involves splitting the BVH into two nodes at each level, then finding the minimal bounding box for each set of positions. We stop splitting while there are still multiple elements in the leaf nodes (4 in this example). (B) We launch a warp of threads for each Gaussian, and compute its AABB. The threads perform a cooperative traversal of the BVH, querying with the AABB, until reaching a leaf node. At this point, threads operate on separate sample positions within the leaf node, compute the Gaussian's influence and value on that position, then atomically add to the sample's output. The threads then continue the traversal, checking every sample position the AABB intersects. After, each sample normalizes its output by dividing its accumulated value by accumulated influence to estimate Equation~\ref{eq:scalar}.  }
    \label{fig:unstructured_method}
    \vspace{-1em}
\end{figure*}

\subsubsection{Densification}
\label{sec:densification}
Training 3D Gaussian models requires adapting the density of Gaussians throughout space via adding Gaussians to under-reconstructed or empty areas (densification) and removing unnecessary Gaussians (pruning). For pruning, we follow Kerbl and Kopanas et al.~\cite{3dgs}, periodically removing Gaussians whose weight falls below 0.005. For densification, we compare three strategies: the gradient heuristic of 3DGS~\cite{3dgs}, the Markov Chain Monte Carlo (MCMC) formulation of Kheradmand et al.~\cite{kheradmand_mcmc_2024}, and our sample-error-based strategy inspired by pixel-error-based methods~\cite{bulo_densification_2024}.

3DGS densifies Gaussians whose view-space positional gradients exceed a threshold; as we have no view space, we use world-space positional gradients instead, and clone/split Gaussians with the largest gradients to meet the same per-densification budget as our approach. MCMC treats the Gaussians as samples of an underlying distribution and recasts densification and pruning as state transitions, relocating low-opacity Gaussians to high-opacity targets chosen by multinomial sampling, then adjusting the relocated Gaussians so the rendered outcome is unchanged. We replace opacity with weight and choose the relocated parameters so that Equation~\ref{eq:scalar} is exactly equal before and after densification, except for a tiny jitter added to new Gaussians' positions to ensure they don't become stuck to those they are cloned from.

Our sample-error-based densification adds new Gaussians at the positions of the $k$ highest-error samples after loss computation, initialized as at model creation with values taken from the ground truth, but weight $0.01$ and scale $\sqrt{0.1}\times$ as large. Rather than always selecting the top $k$, we sample positions without replacement with probability proportional to $\text{error}^\alpha$ (experimentally chosen $\alpha=2.5$ used throughout the evaluation), which prevents the entire densification budget from collapsing onto one region and distributes Gaussians more evenly.
We follow the popular convention of applying densification every 100 iterations of training after iteration 500. Because we target volume compression, we limit densification across training to adding only the number of Gaussians that leads to the desired compression ratio ($|G|$). We also cap the number of Gaussians each event adds so that it takes $\max(3, \lceil\log_{1.5}(|G| / |G_0|)\rceil)$ events to reach the final count $|G|$ from the initial count $|G_0|$. Densification still continues throughout training, though, as pruning opens up slots for new Gaussians to be created. 

\subsubsection{Preventing False Negatives}
\label{sec:continuity}
In contrast to SRNs, a challenge of training VEGGIES models is ensuring the entire domain of a volume's scalar field is represented. Because the set of VEGGIES is explicit, it is possible for there to be positions within the volume where the accumulated influence of the Gaussians does not exceed $\mathrm{T}$. Similarly to samples outside an unstructured volume's geometry (as in Figure~\ref{fig:volumes}), the model's output for the scalar field at these positions is undefined. When positions give defined results in sampling from the original volume but not from the corresponding VEGGIES model, we call the VEGGIES samples false negatives. We show this for an example model in Figure~\ref{fig:fpfn}A. Because the training process optimizes Gaussian parameters based on reconstruction loss of samples it contributed to, it cannot directly solve this error, especially since positions that no Gaussian reaches receive no gradient at all. To overcome this difficulty, we introduce multiple techniques.

First, when computing $G_i(p)$ for a cell, always choosing the cell center as $p$ causes models to overfit to these sample locations. Instead, we select random positions in each cell during training to encourage models to reconstruct space more continuously. Second, when using sample-error-based densification, we include false negative positions as the top priority for adding new VEGGIES. Finally, we introduce a loss term that acts on false negatives and defined samples at risk of becoming false negatives: those whose accumulated influence has dropped close to the threshold $\mathrm{T}$. We apply a logarithmic barrier slightly above $\mathrm{T}$, so the penalty grows rapidly as influence approaches 0, with a small $\epsilon = 0.01$ bounding the penalty and its gradient. For a set of samples defined in the volume ${D}$, we write this at-risk set as ${D}^-$ and the loss term as:
\begin{equation}
\begin{aligned}
{D}^- &= \{ d \in {D} \mid 0 < I_d <  \mathrm{T} + 0.0005 \} \\
L_{FN} &= \lambda_{FN} \cdot \frac{1}{|{D}^-|} \sum_{d \in {D}^-} \max\left(\ln \frac{\mathrm{T} + 0.0005}{I_d + \epsilon \cdot (\mathrm{T} + 0.0005)},\, 0\right)
\end{aligned}
\end{equation}

This term effectively prevents Gaussians from removing their influence on a sample position if it would cause that position to become a false negative. For all models in our evaluation, we set $\lambda_{FN}=10^{-4}$.

\section{Unstructured Volume Reconstruction}
\label{sec:unstructured_method}
Extending the method of Section~\ref{sec:structured_method} to unstructured volumes keeps the same training loop and densification strategies (Section~\ref{sec:structured_training}), but raises two challenges. First, the data points of an unstructured volume are spread irregularly, so the voxel sampler of Section~\ref{sec:structured_implementation} cannot be used; we instead introduce a differentiable sampling algorithm based on a Bounding Volume Hierarchy (BVH) that estimates samples at arbitrary positions (Section~\ref{sec:unstructured_implementation}). Second, the domain of $\phi$ is now defined by the volume's geometry rather than a bounding box, so samples can no longer freely be culled by min and max bounds. This introduces the possibility of false positive samples, which are defined in a VEGGIES model but undefined in the original volume (Figure~\ref{fig:fpfn}B). We address these in Section~\ref{sec:method_fp}. 

\subsection{Unstructured Volume Implementation}
\label{sec:unstructured_implementation}
Our implementation for unstructured volume sampling replaces block-based sampling with BVH construction and traversal for Gaussian-sample intersection testing. We illustrate our method in Figure~\ref{fig:unstructured_method}, consisting of two steps: creating the sample position BVH (Figure~\ref{fig:unstructured_method}A, Section~\ref{sec:bvh}) and using it to estimate samples (Figure~\ref{fig:unstructured_method}B, Section~\ref{sec:unstructured_samples}). Building a BVH over Gaussians and querying with sample positions is less performant due to Gaussian overlap and the number of samples being greater than the number of Gaussians in most use cases. 

\subsubsection{Creating the BVH}
\label{sec:bvh}
In order to sum all Gaussian-sample intersections, we first create a BVH over the input sample positions, allowing us to later query the tree with each Gaussian and efficiently compute output for only meaningful intersections. For BVH creation, we use the CUDA-accelerated radix builder of cuBQL~\cite{cuBQL}, treating each sample position as a 0-volume primitive. Although this can increase the number of unnecessary intersections checked, we allow leaf nodes in the tree to contain many primitives, as it leads to faster BVH creation time and improved parallelism during sample estimation. In our evaluation, we present results using 256 primitives per leaf.

\subsubsection{Estimating Samples}
\label{sec:unstructured_samples}
To query the sample position BVH, we first need to find each Gaussian's AABB using the same preprocessing kernel as Section~\ref{sec:gaussian_bounding}, although now the bounding boxes are created in world space rather than volume blocks. After, the sample estimation kernel is launched with a warp of threads for each Gaussian, which then perform a cooperative traversal of the BVH. This greatly reduces warp divergence, since even nearby Gaussians take different paths during BVH traversal. As the tree was built with many positions per leaf node, when the warp of threads encounters a leaf, each thread can independently compute output for a subset of positions, atomically adding to those positions' global output. This parallelizes the work of each Gaussian, which may have a large number of intersections to compute, while minimizing the heavy cost of redundant BVH traversal that would come from a multiple pass approach towards parallelization. After traversal is completed for each Gaussian, a small kernel is launched to normalize sample output.

After loss is computed, the backward pass to compute Gaussian gradients proceeds much the same as the forward sample estimation. The sample BVH is reused, and a warp of threads is again launched per Gaussian to perform a cooperative traversal. Threads now compute partial gradients from the subset of samples in each leaf they are responsible for, accumulating them into a local sum. After traversal is done, a warp-wide reduction is performed to compute the Gaussian's final gradients. 

\subsection{Preventing False Positives}
\label{sec:method_fp}
A difficulty in extending volume representations to unstructured volumes is the need to encode the volume's domain. Previous work~\cite{liu24_uginr,son2025_mcinr} assumes samples will only be taken inside the unstructured volume, effectively requiring its exterior surface to be stored for any applications that require arbitrary sampling or intersection testing, such as visualization. This greatly limits either the usability of the model, or its maximum compression ratio, since the exterior surface alone is large relative to a compressed model (Table~\ref{tab:unstruct_datasets}). 

Because our method uses the explicit geometry of 3D Gaussians to represent a volume, it naturally defines the domain of the volume as the areas in space within the influence of one or more Gaussians. This allows us to remove the need for storing the geometry of an unstructured volume, as the model encodes the volume's domain directly. In order to train our models for this, we introduce a false positive loss term which penalizes Gaussians for covering positions in space that are not defined by the original volume. Using the same notion of a defined position as above, a position is a false positive when its accumulated influence exceeds $\mathrm{T}$ but it is undefined in the original volume. For the samples undefined in the volume $U$, this false positive term is:
\begin{equation}
\begin{aligned}
L_{FP} &= \lambda_{FP} \cdot \ell \cdot \frac{1}{|{U}|} \sum_{u \in {U}} \max\left(I_u - 0.8\,\mathrm{T},\, 0\right)
\end{aligned}
\end{equation}
where the current L1 error $\ell$ scales the term for better generalization across datasets. This greatly reduces the number of false positives, but leads to an increase in false negatives as Gaussians must now learn to precisely match the spatial domain of the volume rather than extending out past cell boundaries. To counter this, we add a stronger false negative balancing term replacing the one in Section~\ref{sec:continuity}:
 \begin{equation}
\begin{aligned}
L_{FN}' &= \lambda_{FN}' \cdot \ell \cdot \frac{1}{|{D}|} \sum_{d \in {D}} \max\left(1.05\,\mathrm{T} - I_d,\, 0\right)
\end{aligned}
\end{equation}

Throughout our evaluation, we set $\lambda_{FN}'=1200$ and $\lambda_{FP}=12$. With these terms, more VEGGIES are needed to achieve the same reconstruction quality as in the surface storage method, but as shown in Section~\ref{sec:evaluation_unstructured}, this tradeoff is necessary to enable high compression ratios. As an additional optimization, because Gaussians can move far enough outside the volume in early iterations to avoid all training samples, we automatically prune Gaussians who do not have influence on any interior samples above $\mathrm{T}$ between densification events in the first half of training. This mechanism helps ensure that the model does not contain irrelevant Gaussians outside the training area. Note that our false negative and positive losses would not resolve domain encoding for INRs, as these do not report whether a position lies outside the domain. This would require augmentation with another network or external occupancy structure, which has not been explored for unstructured volumes to our knowledge.

\section{Evaluation}
\label{sec:evaluation}
We evaluate the performance of our method in terms of reconstruction quality (PSNR) and training time on sets of both structured (Section~\ref{sec:evaluation_structured}) and unstructured (Section~\ref{sec:evaluation_unstructured}) volume datasets. 
We additionally include a study of different densification strategies (Section~\ref{sec:evaluation_dense}). 
Unless otherwise stated, we use the Adam optimizer~\cite{kingma2017adammethodstochasticoptimization} with experimentally chosen learning rates of 0.005 for values, 0.025 for weights, 0.008 for scaling, 0.005 for rotations, and 0.00016 (decaying toward $1.6\times10^{-6}$ over a 30,000-step schedule) for means. For scale parameters, we apply a soft cap limiting Gaussians to the max of $0.02$ and $1.25\times$ the cube root of the bounding-box volume divided by $|G|$. Because each Gaussian is informed by few samples per iteration, Gaussian parameters can oscillate around an optimum rather than converging to it, so we store an exponential moving average of Gaussian parameters with a decay of $0.995$ for our final model~\cite{polyak1992acceleration}. Models perform best when initialized sparsely so that densification has an adequate budget, so structured models are initialized with $\min(10^{-5}\times|V|, 0.5\times|G|, 20{,}000)$ Gaussians while unstructured models are initialized with $\min(0.5\times|G|, 20{,}000)$ Gaussians, where $|V|$ is the number of data points in the volume (grid points for structured volumes, mesh vertices for unstructured volumes). We then use our probabilistic error-based densification strategy (Section~\ref{sec:densification}) to produce correctly sized models by setting the max number of Gaussians as the target size divided by 48 bytes per Gaussian. We train all of our models for 5,000 iterations. For all training, including comparison works, we use a workstation with an NVIDIA A100 GPU. We report in-memory model sizes for directly queryable models, with no offline file compression applied to them; the sizes we report for SZ3 and TTHRESH are instead the compressed file sizes those methods produce, which favors them in this comparison.

\begin{table*}[t]
\centering
\setlength{\tabcolsep}{3pt}
\resizebox{\textwidth}{!}{%
\begin{tabular}{ll rr rr rr rr rr | rrr rrr}
\toprule
& & \multicolumn{10}{c|}{Directly Queryable Models} & \multicolumn{6}{c}{Traditional Compressors} \\
\cmidrule(lr){3-12} \cmidrule(lr){13-18}
& & \multicolumn{2}{c}{VEGGIES} & \multicolumn{2}{c}{IVNR 20k} & \multicolumn{2}{c}{IVNR 200k} & \multicolumn{2}{c}{AMGSRN} & \multicolumn{2}{c|}{NeurComp} & \multicolumn{3}{c}{SZ3} & \multicolumn{3}{c}{TTHRESH} \\
\cmidrule(lr){3-4} \cmidrule(lr){5-6} \cmidrule(lr){7-8} \cmidrule(lr){9-10} \cmidrule(lr){11-12} \cmidrule(lr){13-15} \cmidrule(lr){16-18}
Dataset & Ratio & PSNR & Time & PSNR & Time & PSNR & Time & PSNR & Time & PSNR & Time & PSNR & Time & Dec. & PSNR & Time & Dec. \\
\midrule
\multirow{3}{*}{Vertebra}
 & 1024$\times$ & \textbf{39.45} & \textbf{8} & 37.29 & \underline{10} & 37.90 & 134 & 38.23 & 120 & \underline{38.65} & 1476 & 36.43 & 2 & 1 & 40.38 & 12 & 3 \\
 & 256$\times$  & \textbf{41.66} & \textbf{8} & 39.20 & \underline{11} & 39.51 & 123 & 39.45 & 108 & \underline{40.63} & 2627 & 42.09 & 2 & 1 & 46.13 & 13 & 4 \\
 & 64$\times$   & \textbf{45.49} & \textbf{10} & 40.98 & \underline{15} & 43.67 & 133 & 41.92 & 120 & \underline{44.06} & 6744 & 57.13 & 2 & 1 & 61.10 & 14 & 4 \\
\midrule
\multirow{3}{*}{Miranda}
 & 1024$\times$ & \textbf{36.36} & \textbf{9} & 31.63 & \underline{12} & 32.73 & 116 & 32.20 & 149 & \underline{36.35} & 37796 & 37.10 & 16 & 9 & 38.95 & 128 & 26 \\
 & 256$\times$  & \textbf{46.45} & \textbf{15} & 37.77 & \underline{17} & 40.76 & 166 & 36.42 & 221 & \underline{44.39} & 99676 & 47.50 & 16 & 9 & 49.41 & 130 & 34 \\
 & 64$\times$   & \textbf{58.27} & \textbf{26} & 39.16 & \underline{36} & 43.22 & 355 & 42.71 & 213 & \underline{55.30} & 327323 & 60.09 & 18 & 10 & 63.45 & 135 & 48 \\
\midrule
\multirow{3}{*}{Chameleon}
 & 1024$\times$ & \textbf{48.62} & \textbf{8} & 41.78 & \underline{11} & 42.98 & 112 & \underline{43.65} & 133 & 43.21 & 40438 & 45.40 & 17 & 9 & 47.83 & 153 & 39 \\
 & 256$\times$  & \textbf{49.93} & \textbf{11} & 46.84 & \underline{16} & \underline{48.40} & 157 & 46.29 & 180 & 43.99 & 104639 & 49.29 & 18 & 9 & 52.72 & 143 & 59 \\
 & 64$\times$   & \textbf{52.03} & \textbf{19} & 48.02 & \underline{31} & 50.50 & 310 & 47.76 & 176 & \underline{51.01} & 353183 & 55.62 & 24 & 11 & 59.84 & 164 & 63 \\
\midrule
\multirow{3}{*}{Richtmyer}
 & 1024$\times$ & \textbf{31.84} & \textbf{11} & 22.65 & \underline{19} & 23.40 & 188 & \underline{27.38} & 840 & --- & --- & 34.11 & 104 & 62 & 34.88 & 1545 & 371 \\
 & 256$\times$  & \textbf{33.39} & \textbf{23} & 23.67 & \underline{48} & 25.45 & 454 & \underline{29.38} & 863 & --- & --- & 44.39 & 115 & 64 & 41.95 & 1403 & 472 \\
 & 64$\times$   & \textbf{33.89} & \textbf{73} & 24.10 & \underline{90} & 26.91 & 859 & \underline{31.12} & 896 & --- & --- & 59.15 & 119 & 72 & 50.98 & 1658 & 659 \\
\bottomrule
\end{tabular}%
}
\caption{Comparison of directly queryable models (VEGGIES at 5k iterations, IVNR at 20k and 200k iterations, AMGSRN at 30k iterations, NeurComp at 50 passes over the volume) and traditional compressors (SZ3, TTHRESH) on reconstruction quality (PSNR) and training / compression time (s). Dec. gives the time (s) to decompress the full volume, which is required for traditional compressors before they can be sampled; directly queryable models require no such step. VEGGIES and INR sizes report in-memory footprints, while SZ3 and TTHRESH report compressed file sizes. \textbf{Bold} indicates best and \underline{underline} second best among directly queryable models. NeurComp entries marked --- did not complete due to unfeasible training times.}
\label{tab:struct_results}
\vspace{-1em}
\end{table*}

\subsection{Structured Volume Evaluation}
\label{sec:evaluation_structured}
\textbf{Experimental Setup} We first present results of our method for several structured volume datasets gathered from the Open SciVis repository~\cite{scivisdata}, including Vertebra ($512^3$, 537 MB), Miranda ($1024^3$, 4.29 GB), Chameleon ($1024\times1024\times1080$, 4.53 GB), and Richtmyer ($2048\times2048\times1920$, 32.21 GB). Before training, these were resampled to floating point and normalized in value range and dimensions to $[0,1]$. Each training step, our method compares samples from the ground truth and VEGGIES model on a $128^3$ grid, where positions are chosen randomly inside each cell. 

For structured volume comparison, we use state-of-the-art INR methods for static volumes along with traditional compressors. We evaluate against the fully CUDA-based implementation of InstantVNR (IVNR)~\cite{wu2024interactivevolumevisualization}, as well as AMGSRN~\cite{wurster25_AMGSRN} with tiny-cuda-nn~\cite{tiny-cuda-nn} for fused CUDA kernels, as these methods have state-of-the-art training time while maintaining quality. In addition, we report results for the fully connected INR NeurComp~\cite{lu_neurcomp_2021} to measure reconstruction quality, though this method is several orders of magnitude slower in training time. NeurComp models diverged in later iterations for Chameleon $256\times$ and $1024\times$, so we report peak PSNR during training for those. For comparison models, we conduct a scan on encoding configurations to achieve target compression ratios, and list these in Tables~\ref{tab:amgsrn_configs}--\ref{tab:neurcomp_configs} of the appendix. We keep configurations as close to those presented in the original work as possible, including the number of iterations used for training. For InstantVNR, we present models trained for both 20k and 200k iterations, where each iteration uses a batch of $65{,}536$ randomly chosen samples; for AMGSRN, we present models trained for 30k iterations with a batch size of $131{,}072$; and for NeurComp, we present models trained for 50 complete passes over the input volume. Because we are reporting in-memory sizes for all directly queryable models, we disable compression-aware loss for AMGSRN according to their recommendation. 

To give baselines for the reconstruction quality measurements reported for directly queryable models, we additionally compare against SZ3~\cite{liang2023sz3} and TTHRESH~\cite{ballesterripoll2020tthresh}, two widely used traditional compressors. SZ3 is an error-bounded prediction and quantization scheme, while TTHRESH is a tensor-decomposition compressor. For all compression ratios, we performed a search over the quality parameter for each method to find a compressed size within $2\%$ of the target. 

\begin{figure}[t]
    \centering
    \includegraphics[width=\linewidth]{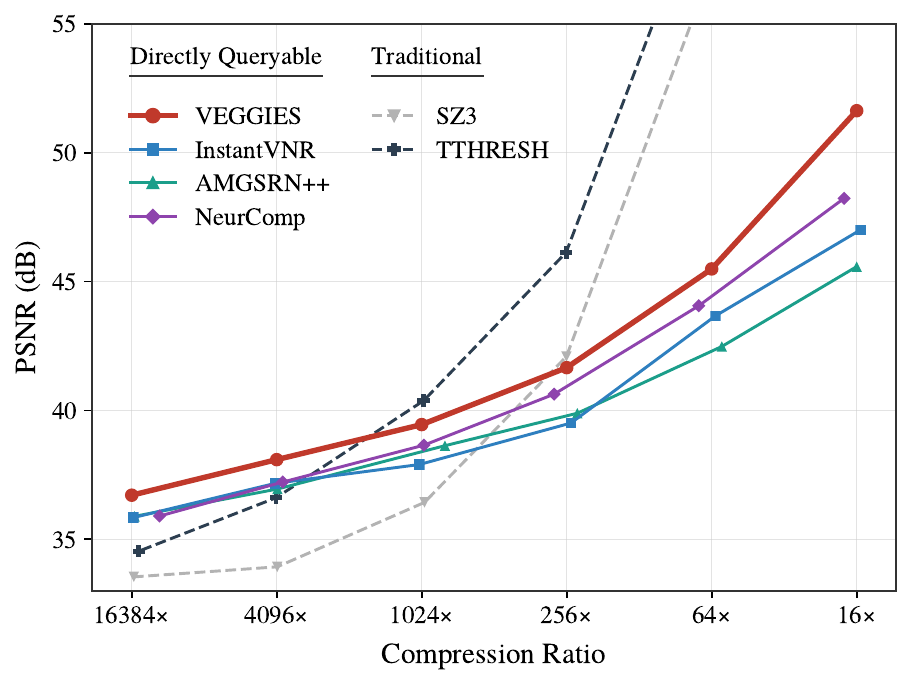}
    \caption{Rate-distortion curves for all structured methods on the Vertebra dataset, evaluated at points 16384, 4096, 1024, 256, 64, and 16. SZ3 and TTHRESH exceed the plotted range, reaching $57.13$ and $61.10$ at $64\times$ and $78.44$ and $86.37$ at $16\times$. Solid lines are directly queryable models, while dashed lines are traditional compressors. }
    \label{fig:rd_curve_struct}
    \vspace{-1.5em}
\end{figure}

\begin{table}[t]
\centering
\setlength{\tabcolsep}{4pt}
\begin{tabular}{@{}l
                S[table-format=4.0]
                S[table-format=3.0]
                S[table-format=5.0]
                |S[table-format=5.0]
                S[table-format=4.0]
                |S[table-format=1.2]
                S[table-format=2.1]@{}}
\toprule
& \multicolumn{3}{c|}{Size Breakdown (MB)}
& \multicolumn{2}{c|}{Size (MB)}
& \multicolumn{2}{c}{Max.\ CR} \\
\cmidrule(lr){2-4}\cmidrule(lr){5-6}\cmidrule(lr){7-8}
Dataset
& {Verts} & {Data} & {Cells} & {Total} & {Surface}
& {Trad.} & {INR} \\
\midrule
RBL        & 9    & 6   & 159   & 174   & 18  & 1.03 & 9.9  \\
Mito       & 12   & 8   & 226   & 246   & 11  & 1.03 & 22.4 \\
SF1        & 30   & 20  & 572   & 622   & 26  & 1.03 & 23.8 \\
Valley     & 439  & 73  & 668   & 1180  & 61  & 1.07 & 19.2 \\
Earthquake & 617  & 206 & 3487  & 4310  & 561 & 1.05 & 7.7  \\
Impact     & 2117 & 706 & 15107 & 17930 & 961 & 1.04 & 18.7 \\
\bottomrule
\end{tabular}
\caption{Unstructured volume datasets and their exterior surfaces, sorted by total size, rounded to nearest MB. Vertices, scalar data, and cells sum to the total (up to rounding). The last two
columns give the maximum real compression ratio achievable by each method. Traditional compressors work only on the scalar data, while INR-based methods must store the exterior surface.}
\label{tab:unstruct_datasets}
\vspace{-1em}
\end{table}

\begin{table*}[t]
\centering
{%
\begin{tabular}{@{}l|c|ccc|ccc|cccc@{}}
\toprule
\multirow{2}{*}{Dataset} & \multirow{2}{*}{Model CR} & \multicolumn{2}{c}{UGINR} & Real & \multicolumn{2}{c}{VEGGIES+S} & Real & \multicolumn{2}{c}{VEGGIES} & Real & \\
& & PSNR & Time (s) & CR & PSNR & Time (s) & CR & PSNR & Time (s) & CR & FP \% \\
\midrule
\multirow{3}{*}{RBL}
 & $64\times$   & 42.83 & 24 & $8.6\times$   & \textbf{103.00} & \underline{13} & $8.6\times$   & \underline{79.32} & \textbf{12} & $64\times$   & 3.72 \\
 & $256\times$  & 30.58 & 23 & $9.5\times$   & \textbf{100.83} & \textbf{11} & $9.5\times$   & \underline{75.94} & \textbf{11} & $256\times$  & 3.64 \\
 & $1024\times$ & 20.72 & 24 & $9.8\times$   & \textbf{95.94} & \underline{12} & $9.8\times$   & \underline{69.89} & \textbf{11} & $1024\times$ & 4.96 \\
\midrule
\multirow{3}{*}{Mito}
 & $64\times$   & 37.84 & 31 & $16.6\times$  & \textbf{92.06} & \textbf{15} & $16.6\times$  & \underline{76.45} & \textbf{15} & $64\times$   & 3.24 \\
 & $256\times$  & 28.80 & 31 & $20.6\times$  & \textbf{90.14} & \textbf{11} & $20.6\times$  & \underline{74.74} & \underline{12} & $256\times$  & 3.41 \\
 & $1024\times$ & 23.15 & 33 & $21.9\times$  & \textbf{85.90} & \textbf{12} & $21.9\times$  & \underline{69.85} & \textbf{12} & $1024\times$ & 4.22 \\
\midrule
\multirow{3}{*}{SF1}
 & $64\times$   & 42.73 & 73 & $17.3\times$  & \textbf{64.35} & \textbf{36} & $17.3\times$  & \underline{61.07} & \underline{38} & $64\times$   & 1.34 \\
 & $256\times$  & 35.55 & 75 & $21.8\times$  & \textbf{60.52} & \textbf{24} & $21.8\times$  & \underline{59.09} & \underline{26} & $256\times$  & 1.39 \\
 & $1024\times$ & 27.05 & 74 & $23.2\times$  & \textbf{56.77} & \textbf{20} & $23.2\times$  & \underline{55.41} & \underline{23} & $1024\times$ & 1.84 \\
\midrule
\multirow{3}{*}{Valley}
 & $64\times$   & 39.20 & 525 & $14.8\times$  & \textbf{55.72} & \underline{63} & $14.8\times$  & \underline{55.13} & \textbf{59} & $64\times$   & 1.19 \\
 & $256\times$  & 37.08 & 508 & $17.9\times$  & \textbf{47.77} & \underline{36}  & $17.9\times$  & \underline{47.31} & \textbf{34}  & $256\times$  & 1.39 \\
 & $1024\times$ & 34.32 & 520 & $18.8\times$  & \textbf{41.05} & \underline{41}  & $18.8\times$  & \underline{40.52} & \textbf{35}  & $1024\times$ & 1.81 \\
\midrule
\multirow{3}{*}{Earthquake}
 & $64\times$   & 51.98 & 1456 & $6.9\times$   & \textbf{71.46} & \underline{141} & $6.9\times$   & \underline{68.42} & \textbf{124} & $64\times$   & 3.17 \\
 & $256\times$  & 50.61 & 1435 & $7.5\times$   & \textbf{68.99} & \underline{62}  & $7.5\times$   & \underline{66.85} & \textbf{57}  & $256\times$  & 3.19 \\
 & $1024\times$ & 45.97 & 1488 & $7.6\times$   & \textbf{63.68} & \underline{47}  & $7.6\times$   & \underline{62.41} & \textbf{42}  & $1024\times$ & 3.56 \\
\midrule
\multirow{3}{*}{Impact}
 & $64\times$   & OOM & OOM & OOM & \textbf{68.37} & \underline{2236} & $14.4\times$  & \underline{67.84} & \textbf{2201} & $64\times$   & 2.00 \\
 & $256\times$  & OOM & OOM & OOM & \textbf{66.53} & \underline{793} & $17.4\times$  & \underline{66.25} & \textbf{787} & $256\times$  & 2.11 \\
 & $1024\times$ & OOM & OOM & OOM & \underline{62.80} & \underline{352}  & $18.3\times$  & \textbf{62.97} & \textbf{309}  & $1024\times$ & 2.12 \\
\bottomrule
\end{tabular}
}
\caption{Comparison of UGINR, our models trained to use the exterior surface for domain checking (VEGGIES+S), and our models trained to encode both scalar fields and volume domain (VEGGIES). Model compression ratios (CR) to the original volume are given, as well as real compression ratios which include exterior surface storage for models which do not encode domain. OOM models were unable to train due to memory constraints. }
\label{tab:evaluation_resultsun}
\vspace{-1em}
\end{table*}

\textbf{Results} We present results for this comparison in Table~\ref{tab:struct_results}, with all models trained for compression ratios of $1024$, $256$, and $64\times$ for each dataset. Against the feature-grid INRs, our method achieves the highest reconstruction quality in every dataset and compression ratio tested. Averaging all twelve benchmarks, VEGGIES reaches a PSNR of 43.12, compared to 38.04 for AMGSRN, 37.95 for IVNR 200k, and 36.09 for IVNR 20k, margins of 5.07, 5.16, and 7.02\,dB. Our method is also the fastest to train of the directly queryable models in all twelve benchmarks, with geometric-mean speedups of $1.42$, $14.4$, and $16.4\times$ over IVNR 20k, IVNR 200k, and AMGSRN. Our method also outperforms the fully connected NeurComp at every benchmark NeurComp completed. Over those nine benchmarks, our models reach a higher average PSNR (46.47 against 44.18) while training $2{,}800\times$ faster (geometric means of 11.7 seconds against 9.2 hours). We were unable to train NeurComp models for Richtmyer at any ratio, as projected training times exceeded 168 hours per model.

The traditional compressors outperform our method on reconstruction quality in most benchmarks, though the gap narrows considerably at high compression ratios. TTHRESH leads in eleven of twelve cells and SZ3 in nine, with the largest margins at $64\times$, where TTHRESH exceeds our PSNR by an average of 11.42\,dB. At $1024\times$ the deficit against TTHRESH falls to 1.44\,dB, and our models lead SZ3 by 0.81\,dB on average at that ratio. To explore this further, we present rate-distortion curves for all methods (using 200k iterations for InstantVNR) on the Vertebra dataset in Figure~\ref{fig:rd_curve_struct}, adding compression ratios $16$, $4096$, and $16384\times$. These curves show that while traditional methods achieve better quality at low compression ratios, directly queryable models outperform them for high compression ratios. In particular, VEGGIES surpasses SZ3 between $256$ and $1024\times$ and TTHRESH between $1024$ and $4096\times$, while also achieving the highest PSNR of any directly queryable method at every compression ratio. We further emphasize that the traditional compressors are not applicable for the same use cases as directly queryable models, because their sampling requires reconstructing the volume first. This costs up to 659 seconds along with the full uncompressed memory footprint of the volume, whereas our models can be sampled directly at any position. Our reported ratios are also uncompressed in-memory footprints, while SZ3 and TTHRESH report entropy-coded file sizes; this offline compression could also be applied to our method but is orthogonal to this work.

\subsection{Unstructured Volume Evaluation}
\label{sec:evaluation_unstructured}
\begin{figure}[t]
    \centering
    \includegraphics[width=\linewidth]{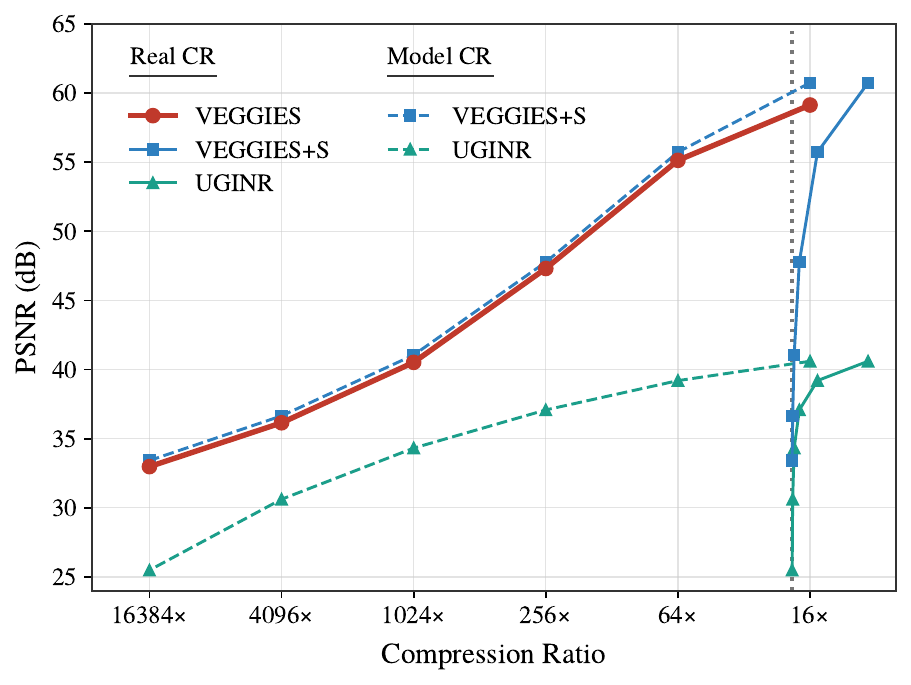}
    \caption{Rate-distortion curves for unstructured methods on the Valley dataset, evaluated at points 16384, 4096, 1024, 256, 64, and 16. Dashed lines are compression ratios using the model size only, while the solid lines include necessary surface storage. The gray dotted line marks the $19.2\times$ ceiling imposed by Valley's 61 MB exterior surface.}
    \label{fig:rd_curve_unstruct}
    \vspace{-1.5em}
\end{figure}

\begin{figure*}[t]
    \centering
    \includegraphics[width=\linewidth]{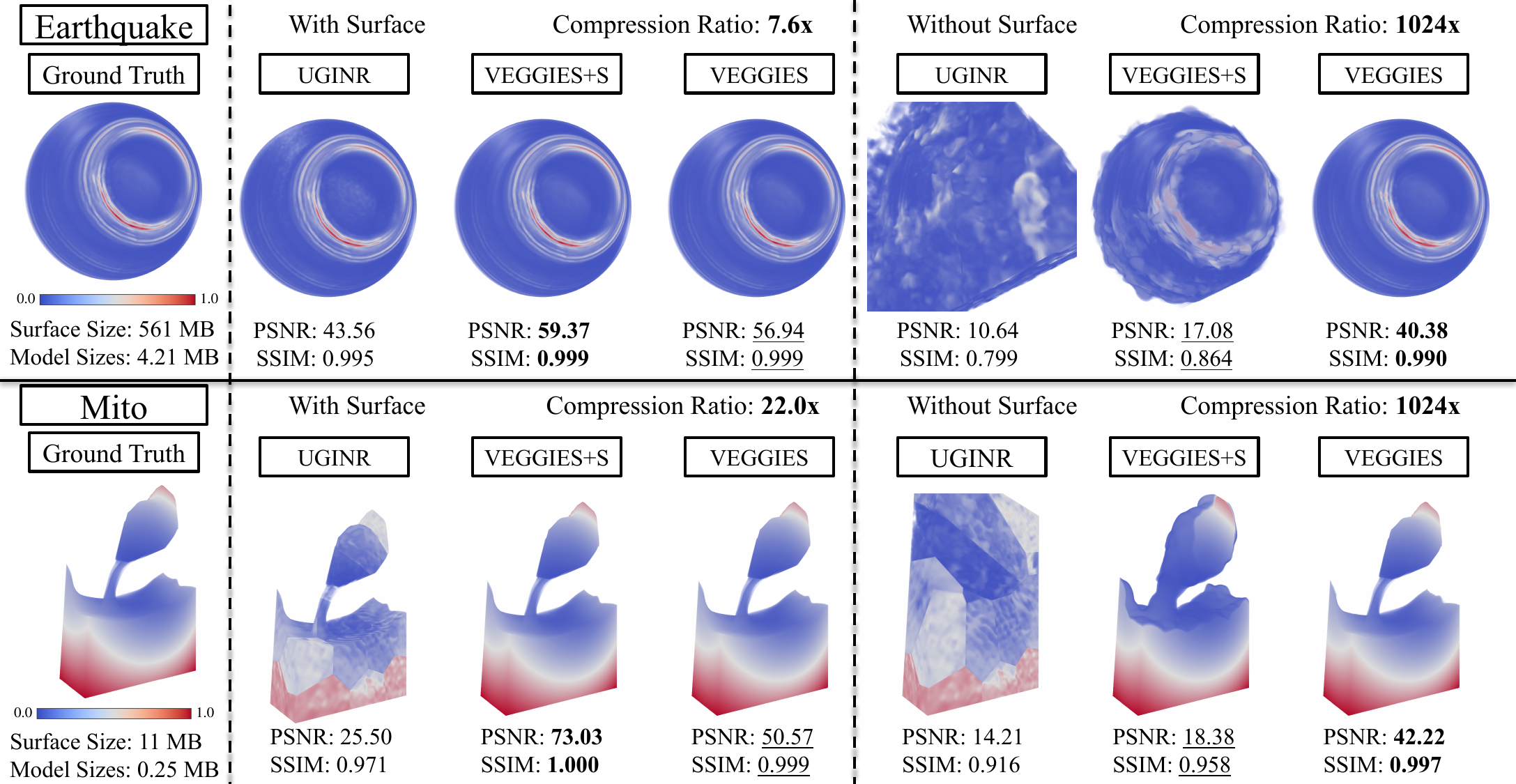}
    \vspace{-1em}
    \caption{Volume rendering using samples taken from UGINR, VEGGIES+S, and VEGGIES models at $1024\times$ model compression ratios for Earthquake and Mito, with and without intersection testing with the datasets' exterior surfaces. VEGGIES models are the only method able to produce accurate samples without intersection testing, removing the need for surface storage and enabling a real $1024\times$ compression ratio.}
    \label{fig:unstructured_renders}
    \vspace{-1em}
\end{figure*}

\textbf{Experimental Setup} We next study the performance of our method on a set of unstructured volumes, all of which come from published sources (RBL, Mito, SF1~\cite{rathke_simd_2015}, Valley~\cite{valley}, Earthquake~\cite{morrical_quickclusters_2023}, Impact~\cite{impact}). We give the size of datasets with regard to unstructured volume vertices, scalar data attached to vertices, cells, total size, and size of the exterior surface of the mesh in Table~\ref{tab:unstruct_datasets}. We additionally include the maximum real compression ratio achievable by traditional and INR-based compression methods for each dataset, assuming that downstream tasks are sampling-oriented, such as volume rendering or isosurface extraction; for these tasks, knowledge of the volumetric field and its domain are necessary, while knowledge of mesh characteristics like its cell connectivity and topology are not. Traditional compression methods can only compress the scalar data, while existing INR-based models effectively compress everything but the exterior surface in these scenarios. This is because they require samples to only be taken inside the unstructured volume's domain, so exterior surfaces must be stored along with the models to support arbitrary sampling. While the memory footprints of these surfaces are much smaller than the datasets themselves, they heavily limit compression, as a surface only one-twentieth the size of the data limits the possible compression ratio to $20\times$.

As can be seen, traditional compression would have almost no effect for these datasets, so we exclude them from our evaluation. For comparing against unstructured volume INRs, we reached out to the authors of MC-INR~\cite{son2025_mcinr}, but their codebase is not available publicly and their paper only evaluates private datasets, so we present results only against UGINR~\cite{liu24_uginr}. We use UGINR's default architecture and training parameters (batch size of $16{,}384$, 75 passes over the volume's data points), except that we cluster on data point positions rather than values so that sampling requires no prior knowledge of a position's value range, and we use 20 clusters as in MC-INR, which we found to outperform UGINR's 75. Time for k-means clustering is excluded from UGINR's reported training time.

We present two kinds of models for our own method: one in which we encode the unstructured volume domain using our explicit model, and one in which we follow the approach of INRs and encode only the scalar field, necessitating exterior surface storage. We refer to these two types of models as VEGGIES and VEGGIES+S (Surface) respectively. For VEGGIES+S models, we follow our structured volume approach by not using false positive loss (Section~\ref{sec:method_fp}), and train using batches of $128^3$ samples taken in the volume. Half of the sample positions are chosen inside uniformly random cells, and half are chosen inside random cells weighted by their value range, effectively assigning training samples according to point density and value variability. To avoid oversampling smaller volumes, we limit the batch size to $\min(128^3, 0.25\times|V|)$. For VEGGIES models, we include false positive loss and use the same sampling method, but replace one-fifth of samples in each batch with random positions just outside the volume, and one-tenth with random positions just inside the volume. These are taken on the boundary surface jittered by a small random amount along the faces' normals either inside or outside. This allows our false positive loss to encourage Gaussians' influence to remain within the domain, using positions that can be arbitrarily close to the volume without being inside. To evaluate the VEGGIES models on domain encoding, our choices of $\lambda_{FN}'$ and $\lambda_{FP}$ ensure models infer a negligible amount of false negatives ($<0.005\%$), so we just compute false positive (FP) percentage by sampling a million undefined points in the same way as above and report what percentage are defined by the model. We list model compression ratios as if the exterior surface does not need to be stored, but also present ``Real CR'', which gives the compression ratio using the sum of the model and surface for UGINR and VEGGIES+S models. For Impact, training for only 5,000 iterations left models far from convergence, so we instead trained for 16,000 iterations, producing much longer training times. 

\textbf{Quantitative Results} We present quantitative results in Table~\ref{tab:evaluation_resultsun}, which show our method achieves the best performance for both reconstruction quality and training time on every dataset UGINR could run on. Averaged over all configurations, VEGGIES+S and VEGGIES improve PSNR over UGINR by \textbf{36.7}\,dB and \textbf{27.6}\,dB, and train over $5.0\times$ faster (geometric mean). Neither wins by less than 6.20\,dB or $1.85\times$ on any configuration. UGINR was unable to run on the large Impact dataset.

By allowing some lossiness on reconstructing the unstructured volume domain, VEGGIES models can drastically increase real compression ratios, reaching $18\times$ higher than VEGGIES+S across the configurations in Table~\ref{tab:evaluation_resultsun} (geometric mean). The reconstruction quality lost from VEGGIES+S to VEGGIES is small to moderate on the larger datasets (from 0--3.3\,dB on SF1, Valley, Earthquake, and Impact) but significant on RBL and Mito, whose very simple scalar fields are defined on complex geometry, causing domain encoding to consume most of the model's capacity. Even at that cost, VEGGIES still outperforms UGINR on both PSNR and training time for every benchmark. Figure~\ref{fig:rd_curve_unstruct} presents rate-distortion curves with real and model compression ratios for all methods on the Valley dataset, adding points for $16$, $4096$, and $16384\times$. Measured by model CR, VEGGIES+S is slightly ahead of VEGGIES at every rate, but once the 61\,MB exterior surface is included, VEGGIES+S and UGINR hit a hard $19.2\times$ ceiling for real compression ratio. VEGGIES has no such ceiling, reaching real compression up to $16384\times$ while maintaining a PSNR above 30 dB. 

\textbf{Qualitative Results} We additionally include a visual comparison of UGINR, VEGGIES+S, and VEGGIES models by integrating them into the NerfAcc~\cite{li2023nerfacc}-accelerated volume renderer of AMGSRN~\cite{wurster25_AMGSRN}. We extend their renderer with the ability to discard samples outside an unstructured volume's domain by intersection testing with its exterior surface. We present rendering results from each method's $1024\times$ Mito and Earthquake models in Figure~\ref{fig:unstructured_renders}, both with and without surface testing. Each render uses $1{,}024$ samples per pixel, at a resolution of $1{,}024\times1{,}024$, using the matplotlib ``coolwarm'' colormap and linear opacity map. We render ground truth images using samples taken from the original volume, and compute image-based PSNR and SSIM for model renders. For both datasets, VEGGIES models alone achieve reasonable quality when rendering without surface testing. This validates our method's ability to represent an unstructured volume without any mesh storage, enabling previously impossible compression ratios. 


\begin{table}[t]
\centering
\setlength{\tabcolsep}{4pt}
\begin{tabular}{ll r|r|r|r}
\toprule
Dataset & Metric & 3DGS & MCMC & Error & Prob.\ Error \\
\midrule
\multirow{2}{*}{Miranda}
  & PSNR & 42.39 & 41.52 & \underline{45.79} & \textbf{46.45} \\
  & Time & 16 & 36 & \textbf{14} & \underline{15} \\
\midrule
\multirow{2}{*}{Chameleon}
  & PSNR & 49.29 & 48.48 & \underline{49.83} & \textbf{49.93} \\
  & Time & 17 & 36 & \textbf{10} & \underline{11} \\
\midrule
\multirow{2}{*}{Valley+S}
  & PSNR & 46.85 & 45.55 & \underline{47.51} & \textbf{47.77} \\
  & Time & 45    & 41    & \textbf{35}    & \underline{36} \\
\midrule
\multirow{3}{*}{Valley}
  & PSNR & 45.30 & 45.06 & \underline{47.04} & \textbf{47.31} \\
  & Time & 46    & 42    & \textbf{33}    & \underline{34} \\
  & FP\% & \textbf{1.19} & \underline{1.32} & 1.42 & 1.39 \\
\midrule
\multirow{2}{*}{Earthquake+S}
  & PSNR & 63.78 & 62.86 & \underline{66.50} & \textbf{68.99} \\
  & Time & 107   & 113   & \textbf{57} & \underline{62} \\
\midrule
\multirow{3}{*}{Earthquake}
  & PSNR & 62.34 & 60.35 & \underline{65.61} & \textbf{66.85} \\
  & Time & 90 & 100 & \textbf{53} & \underline{57} \\
  & FP\% & \textbf{2.86} & 3.65 & 3.23 & \underline{3.19} \\
\bottomrule
\end{tabular}
\caption{Comparison of 3DGS, MCMC, and our error-based and probabilistic
error-based densification strategies on reconstruction quality (PSNR), training
time (in seconds), and false positive rate.}
\label{tab:dense_results}
\vspace{-1em}
\end{table}

\subsection{Densification}
\label{sec:evaluation_dense}
In this section, we compare the three densification strategies outlined in Section~\ref{sec:densification} on two structured (Miranda, Chameleon) and two unstructured (Valley, Earthquake) datasets. For models using MCMC, we tuned its regularization and noise rate terms per dataset for best PSNR. We compare by creating models at $256\times$ compression, trained using the various densification strategies. For unstructured volumes, we present both VEGGIES+S and VEGGIES models. We report results in Table~\ref{tab:dense_results}. These results show that our error-based densification strategies lead to the highest reconstruction quality for every dataset, by an average margin of 2.06\,dB for error-based and 2.89\,dB for probabilistic error-based over the stronger of the two baselines for each dataset. 


\section{Conclusion}
\label{sec:limitations}
We have proposed a method to effectively train an explicit 3D Gaussian model for structured and unstructured volume data compression. With our efficient sampling algorithms, loss functions which improve domain encoding, and probabilistic error-based densification strategy, our representation reaches higher reconstruction quality than every INR tested, while training faster than every INR tested and orders of magnitude faster than the most accurate. Due to our model's unique ability to encode an unstructured volume's domain, we remove previous work's extremely limiting hard cap on compression ratios.

Our method has several potential areas for future work, including a natural extension to time-varying and multivariate datasets. Notably, we do not implement any of the wealth of literature on compressing 3DGS models for storage on disk~\cite{papantonakis_compressedgs_2024, niedermayr_compressedgs_2024, fan_lightgaussian_2025}. Another improvement could be development of a loss function based on signed distance fields (SDF) for unstructured volume domains, although our initial experimentation found SDF evaluation to be prohibitively expensive. Finally, while our models support arbitrary visualization tasks, we leave a thorough investigation of accuracy for specific operations to future work. Implementing a rendering engine designed specifically to sample from our models could lead to interesting developments, and recent work in image-trained 3D Gaussian models has shown ray tracing 3D Gaussians to be a promising extension~\cite{loccoz20243dgrt}.


\bibliographystyle{abbrv-doi-hyperref}

\bibliography{template}

\appendix 

\section{Model Configurations}
In this section, we list the network configurations used for the structured dataset comparison models discussed in Section~\ref{sec:evaluation_structured}. We list the configurations for AMGSRN models in Table~\ref{tab:amgsrn_configs}, for InstantVNR models in Table~\ref{tab:instantvnr_configs}, and for NeurComp models in Table~\ref{tab:neurcomp_configs}.

\begin{table}[h]
\centering
\begin{tabular}{llccc}
\toprule
Dataset & & \multicolumn{3}{c}{Compression ratio} \\
\midrule
\multirow{6}{*}{Vertebra}
 & Ratio                & $1024\times$ & $256\times$ & $64\times$ \\
 & Features              & 2       & 1       & 4       \\
 & Grids                 & 16      & 16      & 16      \\
 & Grid Shape           & $16^3$  & $32^3$  & $32^3$  \\
 & Layers                & 2       & 2       & 2       \\
 & Nodes per Layer      & 64      & 64      & 64      \\
\midrule
\multirow{6}{*}{Vertebra (cont.)}
 & Ratio                & $16384\times$ & $4096\times$ & $16\times$ \\
 & Features              & 2       & 1       & 2       \\
 & Grids                 & 16      & 16      & 16      \\
 & Grid Shape           & $3^3$   & $11^3$  & $63^3$  \\
 & Layers                & 2       & 2       & 2       \\
 & Nodes per Layer      & 64      & 64      & 64      \\
\midrule
\multirow{6}{*}{Miranda}
 & Ratio                & $1024\times$ & $256\times$ & $64\times$ \\
 & Features              & 1      & 4      & 2      \\
 & Grids                 & 32     & 32     & 32     \\
 & Grid Shape           & $32^3$ & $32^3$ & $64^3$ \\
 & Layers                & 2      & 2      & 2      \\
 & Nodes per Layer      & 64     & 64     & 64     \\
\midrule
\multirow{6}{*}{Chameleon}
 & Ratio                & $1024\times$ & $256\times$ & $64\times$ \\
 & Features              & 1      & 4      & 2      \\
 & Grids                 & 32     & 32     & 32     \\
 & Grid Shape           & $32^3$ & $32^3$ & $64^3$ \\
 & Layers                & 2      & 2      & 2      \\
 & Nodes per Layer      & 64     & 64     & 64     \\
\midrule
\multirow{6}{*}{Richtmyer}
 & Ratio                & $1024\times$ & $256\times$ & $64\times$ \\
 & Features              & 1       & 4       & 2        \\
 & Grids                 & 32      & 32      & 32       \\
 & Grid Shape           & $64^3$  & $64^3$  & $128^3$  \\
 & Layers                & 2       & 2       & 2        \\
 & Nodes per Layer      & 64      & 64      & 64       \\
\bottomrule
\end{tabular}
\caption{AMGSRN configurations for structured datasets. }
\label{tab:amgsrn_configs}
\end{table}

\begin{table}[h]
\centering
\resizebox{\ifdim\width>\columnwidth\columnwidth\else\width\fi}{!}{%
\begin{tabular}{llccc}
\toprule
Dataset & & \multicolumn{3}{c}{Compression ratio} \\
\midrule
\multirow{7}{*}{Vertebra}
 & Ratio                     & $1024\times$ & $256\times$ & $64\times$ \\
 & Neurons                   & 64 & 64 & 64 \\
 & Hidden Layers             & 2  & 2  & 2  \\
 & Levels                     & 10 & 8  & 8  \\
 & Features per Level        & 4  & 1  & 4  \\
 & Log$_2$ Hashmap Size      & 13 & 18 & 18 \\
 & Base Resolution           & 4  & 4  & 4  \\
\midrule
\multirow{7}{*}{Vertebra (cont.)}
 & Ratio                     & $16384\times$ & $4096\times$ & $16\times$ \\
 & Neurons                   & 64 & 64 & 64 \\
 & Hidden Layers             & 2  & 2  & 2  \\
 & Levels                     & 3  & 6  & 9  \\
 & Features per Level        & 4  & 2  & 2  \\
 & Log$_2$ Hashmap Size      & 11 & 13 & 21 \\
 & Base Resolution           & 4  & 4  & 4  \\
\midrule
\multirow{7}{*}{Miranda}
 & Ratio                     & $1024\times$ & $256\times$ & $64\times$ \\
 & Neurons                   & 64 & 64 & 64 \\
 & Hidden Layers             & 3  & 3  & 3  \\
 & Levels                     & 5  & 6  & 12 \\
 & Features per Level        & 4  & 8  & 8  \\
 & Log$_2$ Hashmap Size      & 19 & 19 & 19 \\
 & Base Resolution           & 4  & 4  & 4  \\
\midrule
\multirow{7}{*}{Chameleon}
 & Ratio                     & $1024\times$ & $256\times$ & $64\times$ \\
 & Neurons                   & 64 & 64 & 64 \\
 & Hidden Layers             & 3  & 3  & 3  \\
 & Levels                     & 5  & 8  & 9  \\
 & Features per Level        & 4  & 4  & 8  \\
 & Log$_2$ Hashmap Size      & 19 & 19 & 20 \\
 & Base Resolution           & 4  & 4  & 4  \\
\midrule
\multirow{7}{*}{Richtmyer}
 & Ratio                     & $1024\times$ & $256\times$ & $64\times$ \\
 & Neurons                   & 64 & 64 & 64 \\
 & Hidden Layers             & 5  & 5  & 5  \\
 & Levels                     & 6  & 12 & 13 \\
 & Features per Level        & 8  & 8  & 8  \\
 & Log$_2$ Hashmap Size      & 19 & 20 & 22 \\
 & Base Resolution           & 4  & 4  & 4  \\
\bottomrule
\end{tabular}%
}
\caption{InstantVNR configurations for structured datasets.}
\label{tab:instantvnr_configs}
\end{table}

\begin{table}[h]
\centering
\resizebox{\ifdim\width>\columnwidth\columnwidth\else\width\fi}{!}{%
\begin{tabular}{llccc}
\toprule
Dataset & & \multicolumn{3}{c}{Compression ratio} \\
\midrule
\multirow{3}{*}{Vertebra}
 & Ratio                & $1024\times$ & $256\times$ & $64\times$ \\
 & Neurons per Layer    & 96 & 181 & 363 \\
 & Hidden Layers        & 8  & 8   & 8   \\
\midrule
\multirow{3}{*}{Vertebra (cont.)}
 & Ratio                & $16384\times$ & $4096\times$ & $16\times$ \\
 & Neurons per Layer    & 23 & 47 & 728 \\
 & Hidden Layers        & 8  & 8  & 8   \\
\midrule
\multirow{3}{*}{Miranda}
 & Ratio                & $1024\times$ & $256\times$ & $64\times$ \\
 & Neurons per Layer    & 272 & 514 & 1030 \\
 & Hidden Layers        & 8   & 8   & 8    \\
\midrule
\multirow{3}{*}{Chameleon}
 & Ratio                & $1024\times$ & $256\times$ & $64\times$ \\
 & Neurons per Layer    & 280 & 528 & 1058 \\
 & Hidden Layers        & 8   & 8   & 8    \\
\bottomrule
\end{tabular}%
}
\caption{NeurComp configurations for structured datasets. All models use residual SIREN layers with $\omega_0=30$ and 50 passes over the volume, with the learning rate decayed by $0.2\times$ every 20 passes. The initial learning rate is NeurComp's $100/P$ for $P$ network parameters, capped at $5\times10^{-5}$. The batch size was $65{,}536$ for all models but Miranda and Chameleon $64\times$, as GPU memory limits required using $32{,}768$ for twice as many steps instead. }
\label{tab:neurcomp_configs}
\end{table}

\section{Testing Image-Trained VEG}
\label{sec:veg_failure}

\begin{figure}[h]
    \centering
    \includegraphics[width=\linewidth]{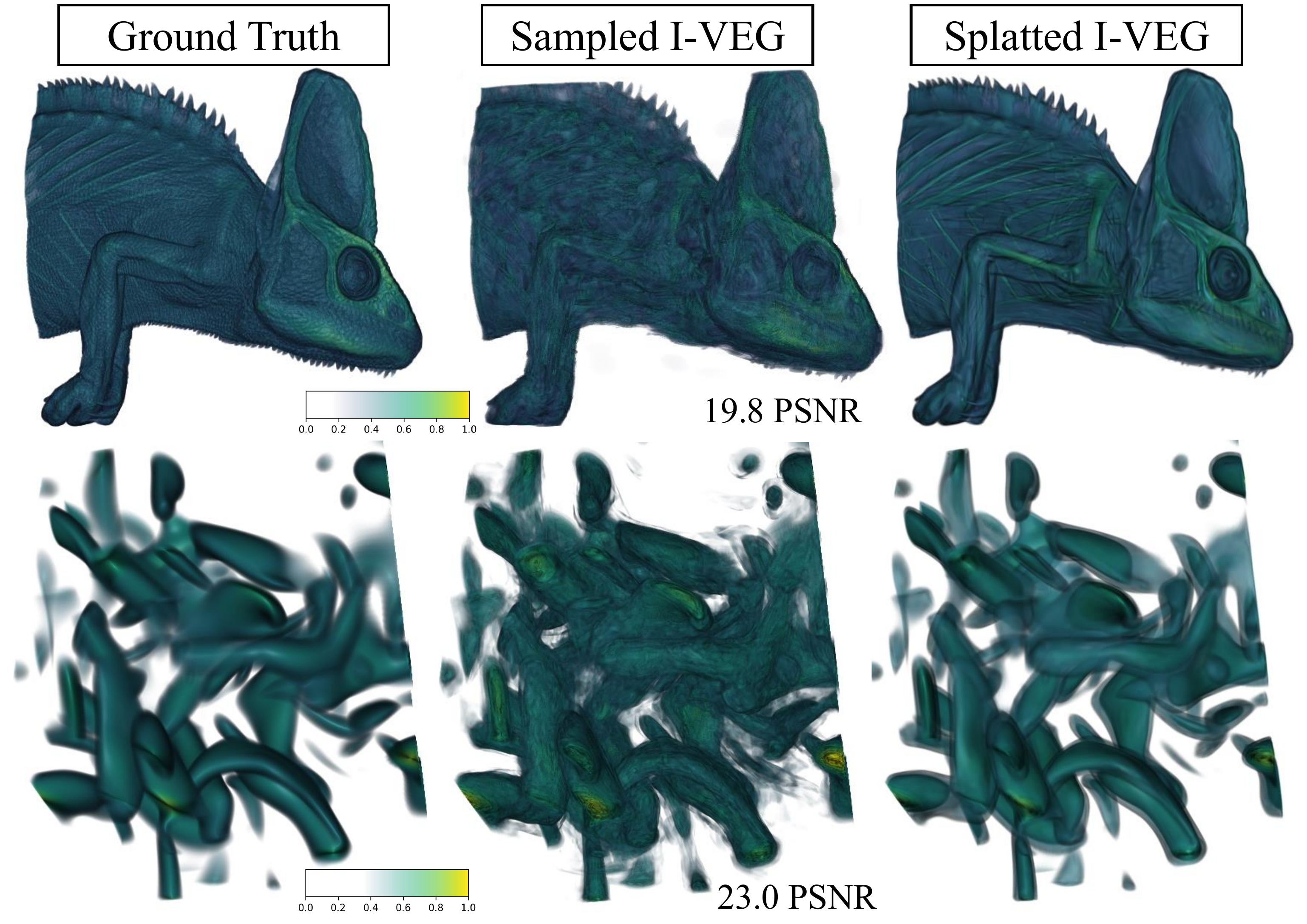}
    \caption{``Chameleon'' and ``Vortex'' I-VEG models from Dyken et al.~\cite{veg_2025}, with splat-based renders using their method. To assess performance for volume reconstruction, we sample each model to a $512^3$ grid. We present volumetric PSNR comparing sampled volumes to the ground truth, as well as renders using PyVista.}
    \label{fig:sampledVEG}
    \vspace{-1.5em}
\end{figure}

To motivate our development of a world-space VEGGIES training method for volume compression, we evaluate sampling of models produced by image-based training. We obtained the ``Chameleon'' and ``Vortex'' I-VEG models from Dyken et al.~\cite{veg_2025}, and sampled them onto a $512^3$ regular grid using our inference method (Equation~\ref{eq:scalar}). Figure~\ref{fig:sampledVEG} presents the resulting volumes rendered with PyVista, alongside reconstruction quality as volumetric PSNR. The I-VEG models give low PSNR values of 19.8 and 23.0\,dB, respectively, showing that models learned through image-space training cannot be immediately used for accurate sampling. Basic interventions of scaling Gaussian extents and weights by different factors did not improve reconstruction quality.

\end{document}